\documentclass[conference]{IEEEtran}
\IEEEoverridecommandlockouts
\usepackage{cite}
\usepackage{amsmath,amssymb,amsfonts}
\usepackage{algorithmic}
\usepackage{graphicx}
\usepackage{textcomp}
\usepackage{xcolor}
\def\BibTeX{{\rm B\kern-.05em{\sc i\kern-.025em b}\kern-.08em
    T\kern-.1667em\lower.7ex\hbox{E}\kern-.125emX}}

\usepackage{tikz}
\usetikzlibrary{shapes.geometric, positioning, calc, arrows.meta}
\usepackage{multirow}
\usepackage{algorithm2e}
\RestyleAlgo{ruled}
\usepackage{booktabs}
\usepackage{tabularx}

\tikzset{
  -|-/.style={
    to path={
      (\tikztostart) -| ($(\tikztostart)!#1!(\tikztotarget)$) |- (\tikztotarget)
      \tikztonodes
    }
  },
  -|-/.default=0.5,
  |-|/.style={
    to path={
      (\tikztostart) |- ($(\tikztostart)!#1!(\tikztotarget)$) -| (\tikztotarget)
      \tikztonodes
    }
  },
  |-|/.default=0.5,
}

\begin{document}

\title{Explainable Metric Learning for Deflating Data Bias}

\author{Emma Andrews\\
University of Florida\\
{\tt\small e.andrews@ufl.edu}
\and
Prabhat Mishra\\
University of Florida\\
{\tt\small prabhat@ufl.edu}
}

\maketitle

\begin{abstract}
Image classification is an essential part of computer vision which assigns a given input image to a specific category based on the similarity evaluation within given criteria. While promising classifiers can be obtained through deep learning models, these approaches lack explainability, where the classification results are hard to interpret in a human-understandable way. In this paper, we present an explainable metric learning framework, which constructs hierarchical levels of semantic segments of an image for better interpretability. The key methodology involves a bottom-up learning strategy, starting by training the local metric learning model for the individual segments and then combining segments to compose comprehensive metrics in a tree. Specifically, our approach enables a more human-understandable similarity measurement between two images based on the semantic segments within it, which can be utilized to generate new samples to reduce bias in a training dataset. Extensive experimental evaluation demonstrates that the proposed approach can drastically improve model accuracy compared with state-of-the-art methods.
\end{abstract}


\section{Introduction}
\label{sec:intro}
In the computer vision domain, machine learning (ML) models are widely applied for pattern classification or facial recognition, which often involves comparing the similarity between images. Deep metric learning has been proposed in this field as an efficient way of measuring similarity. 
Traditional deep metric learning focuses on mapping input images into feature space where the distances between samples reflect their semantic or perceptual similarities. The goal is to learn an optimal distance metric through their pixel values. However, the pixel-based similarity results cannot be interpreted in a human-understandable way since humans tend to view an image as a composition of important regions or meaningful segments. For example, a human is likely to compare the eyes/noses/ears between two facial images as a singular entity instead of comparing each pixel as a computer would in traditional machine learning.



Although a saliency map-based approach~\cite{zhang2022attributable} could effectively emphasize crucial regions as determined by a machine learning model in an image, it faces several practical limitations. (1) For example, specific domains, such as facial recognition, prioritize individual features that may not be captured by a model. (2) Moreover, employing a segmentation-based approach becomes imperative to recognize features that may not be highlighted by a saliency map. This ensures that similarity scores, obtained through deep metric learning, accurately reflect the importance of diverse features in the recognition process. In other words, a segmentation-based approach is more human-understandable. (3) A saliency map-based approach is expected to introduce significant overhead since it needs to compute regions for each image. In contrast, a segmentation-based approach is expected to be lightweight since the segments need to be generated once.

\begin{figure}[h]
\vspace{-0.2in}
    \centering
    \begin{tikzpicture}[
        block/.style={
        draw,
        fill=white,
        rectangle, 
        minimum width={width("nose")},
        align=center,
        minimum height=0.5cm,
        font=\small}]
        \node[block] (l1) at (0,0) {full image};
        \node[block] (l2) at (-1, -1) {neck, hair};
        \node[block] (l3) at (1, -1) {nose, eyes};
        \node[block] (l4) at (-1.5, -2) {neck};
        \node[block] (l5) at (-0.5, -2) {hair};
        \node[block] (l6) at (0.5, -2) {nose};
        \node[block] (l7) at (1.5, -2) {eyes};
        \draw[->] (l1.south) -- (l2.north);
        \draw[->] (l1.south) -- (l3.north);
        \draw[->] (l2.south) -- (l4.north);
        \draw[->] (l2.south) -- (l5.north);
        \draw[->] (l3.south) -- (l6.north);
        \draw[->] (l3.south) -- (l7.north);
    \end{tikzpicture}
    \caption{An overview of our proposed framework. A model is trained at each individual and combined segment until the fully reconstructed image is reached. Example segments are from CelebA~\cite{liu2015deep}.}
    \label{fig:proposed}
\end{figure}
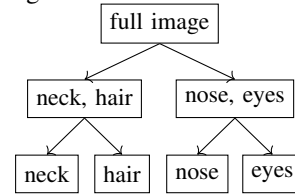

In this paper, we propose a new framework, Hierarchical Explainable Metric Learning (HEML) for constructing an explainable similarity metric using semantic segmentation. The basic idea is to break down the images into their semantic segments, train a metric learning model for each segment, and progressively combine the segments to reconstruct the original image. At each stage of combination, separate deep metric learning models are trained to compute the combinational similarity metrics among individual segments within an image. Models trained at lower levels contribute to the training of higher levels, iteratively refining the weights in a bottom-up manner throughout the image reconstruction process.
Figure~\ref{fig:proposed} shows a simple example demonstration of our proposed approach with the segment combination.

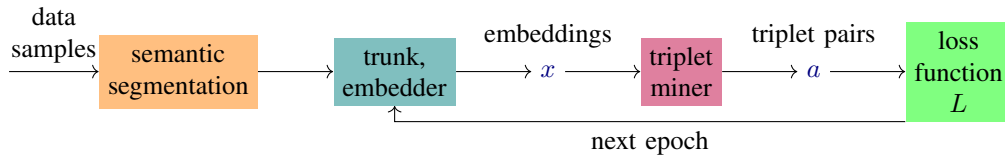
\begin{figure*}[h]
    \centering
    \begin{tikzpicture}
    
      \node[fill=orange!50, align=center] (l1) {semantic\\segmentation};
      \node[fill=teal!50, right=of l1, align=center] (l2) {trunk,\\embedder};
      \node[blue!50!black, right=of l2, label={above:embeddings}] (act1) {$x$};
      \node[fill=purple!50, right=of act1, align=center] (l3) {triplet\\miner};
      \node[blue!50!black, right=of l3, label={above:triplet pairs}] (act2) {$a$};
      \node[fill=green!50, right=of act2, align=center] (l4) {loss\\function\\$L$};
    
      \draw[->] (l1) -- (l2);
      \draw[<-] (l1) -- ++(-2.25,0) node[above, pos=0.5, align=center] {data\\samples};
      \draw[->] (l2) -- (act1);
      \draw[->] (act1) -- (l3);
      \draw[->] (l3) -- (act2);
      \draw[->] (act2) -- (l4);
      \draw[<-] (l2.south) |- (l4.south west) node[below, pos=0.75, align=center] {next epoch};
    
    \end{tikzpicture}
    \caption{Overview of our proposed hierarchical explainable metric learning framework. It consists of three major tasks: semantic segmentation, bottom-up metric learning, and metric tree construction.}
    \vspace{-0.1in}
    \label{fig:overview}
\end{figure*}

The rest of the paper is organized as follows. Section~\ref{sec:relwork} examines related work and establishes necessary background information. Section~\ref{sec:method} explains the HEML methodology in detail. Section~\ref{sec:result} presents experimental results. Finally, we conclude this paper in Section~\ref{sec:conclusion}.

\section{Background and Related Work}
\label{sec:relwork}

In this section, we first provide an overview of explainable AI. Next, we discuss existing efforts related to explainable metric learning and discuss their limitations to highlight the need for the proposed work.

\subsection{Explainabile Artificial Intelligence}
Explainable AI focuses on providing human-understandable explanations of the steps that a machine learning model took to reach a certain decision or outcome \cite{adadi2018peeking, barredoarrieta2020explainable, dosilovic2018explainable}. For images specifically, explainable AI can be used to determine the importance of certain pixels within an image that contributed the most to a model's decision. Popular explainable AI methods for images include saliency maps and integrated gradients.

Saliency maps highlight important pixels that a machine learning model uses in its decision-making, often resulting in a heatmap \cite{molnar2022interpretable}. The heatmap typically features a concentration of the most important pixels in a single area of the image. Integrated gradients also generate the most important pixels, however, it does so based on the gradients calculated throughout training \cite{sundararajan2017axiomatic, qi2020visualizing}. Compared to saliency maps, integrated gradients typically result in more specific highlighting of pixels with less emphasis on regions.

XRAI attempts to expand integrated gradients into more concrete regions~\cite{kapishnikov2019xrai}. XRAI optimizes integrated gradients to avoid issues with black pixels, then extracts the resulting important pixels into a region. These regions are often better aligned with what a human would expect to be the most important region within an image for the current task.

\subsection{Explainable Deep Metric Learning}
Deep metric learning aims to decrease the distance between similar samples and increase the distance between dissimilar samples \cite{kaya2019deep}. The distances between samples are calculated between an embedding representation of the sample. This embedding is the output from a convolutional neural network (CNN), such as ResNet. Often, this resulting output embedding is further reduced in dimensionality by a simple multi-layer perception (MLP) model.
Much of the advancement within deep metric learning is for loss function optimizations. Triplet margin loss was proposed by Schroff et al. to use triplet pairs to identify positive and negative pairs for easier similarity optimizations \cite{schroff2015facenet}. ArcFace and Sub-center ArcFace are optimized for running on facial images \cite{deng2018arcface, deng2020subcenter}. 

Additional advancement comes from deep metric learning strategies. Adversarial metric learning uses a generator and discriminator to create hard sample pairs to learn the embedding space \cite{chen2018adversarial}. Deep causal metric learning trains distance metrics by utilizing causal inference, establishing the cause-and-effect relationship between embeddings \cite{deng2022deep}. While specialized loss functions and strategies can help learn a better embedding representation, it does not lend itself to explainability by default. Extra steps are necessary to explain the distance and similarities between samples.

Deep Interpretable Metric Learning (DIML) is the first known work for adding explainability to deep metric learning \cite{zhao2021interpretable}. Instead of using the normal embeddings from the embedder model, these embeddings are further reduced into structural embeddings before running distance or similarity metrics.
DIML is expanded with Attributable Visual Similarity Learning (AVSL). AVSL is an approach to explainable similarity learning designed to determine the similarity between images based on their saliency maps \cite{zhang2022attributable}. 

While existing approaches are promising, their applicability is limited for various reasons. First, the saliency maps are used to gather the pixels at each layer and their importance to its overall classification. Since we need to compute a saliency map for each image, it can be a memory bottleneck for many application scenarios. 
Furthermore, the features extracted to be broken down are obtained by the feature extraction of a CNN. Since the CNN is doing the feature extraction, it is again reliant on a machine to determine what areas of an image are a specified segment. This feature can be in the latent space and therefore may not be human-explainable. To the best of our knowledge, our approach is the first attempt at creating a lightweight solution for human-understandable metric learning.


\section{Hierarchical Explainable Metric Learning}
\label{sec:method}
We propose a new framework called Hierarchical Explainable Metric Learning (HEML) for composing an explainable tree of semantic segmentation similarities for two images. This framework consists of three major parts: semantic segmentation, bottom-up deep metric learning, and metric tree construction. Figure \ref{fig:overview} shows an overview of our methodology. We first feed our data samples into a SegFormer for semantic segmentation. These segmented images are then used as samples to train individual trunk and embedder models as part of the main deep metric learning training loop. From the output of the models, we use a triplet miner to find hard pairs to target in the loss function.


\subsection{Semantic Segmentation}
\label{sec:ss}

\begin{figure}[h]
    \centering
    \includegraphics[scale=0.25]{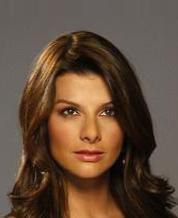}
    \includegraphics[scale=0.25]{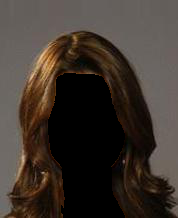}
    \includegraphics[scale=0.25]{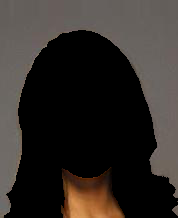}
    \includegraphics[scale=0.25]{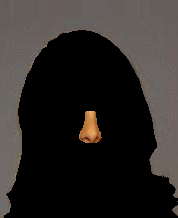}
    \includegraphics[scale=0.25]{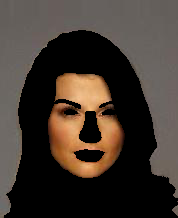}
    \caption{\textbf{Left image:} Original sample image from CelebA. \textbf{Remaining images:} Example segments combined with background. From left to right: hair, neck, nose, and skin segments.}
    \label{fig:sample}
\end{figure}

Image segmentation is the process of breaking an image into meaningful regions. These regions are often major areas of importance within an image, especially to a human. For example, given an image of a house, semantic regions that can be extracted are that of doors, windows, walls, and roofs. 
SegFormers are the most popular machine learning model for performing automatic image segmentation. These models are a specific form of the transformer model with the task of semantic segmentation of an image based on specific labels \cite{vaswani2017attention, xie2021segformer, parmar2018image, guo2018review}. Figure \ref{fig:sample} showcases sample segments from a CelebA image sample.


Once the individual segments have been created by the SegFormer, two individual segments are combined together. This is done for all individual segment pairs, which are then combined together with another combined image. This process repeats until the entire image has been reconstructed. Our image segmentation process preserves each individual segmented feature as an image, so different sets of features can be combined without having to repeat the segmentation pipeline. This pixel-level transformation provides a detailed and easily reconfigurable training dataset for the bottom-up metric learning stage.

\subsection{Bottom-Up Metric Learning}
\label{sec:bup}
Let $D$ be the image dataset as a collection of instances, each can be represented as a feature vector $x_i$ in the feature space $X$. Formally, $D = \{(x_1,y_1), (x_2,y_2), ..., (x_N,y_N)\}$, where $N$ is the number of images, and $y_i$ is the corresponding label for $x_i$. 

In the context of metric learning, a metric $d$ is defined as function $d: X \times X \rightarrow \mathbb{R}^+$ to measure the similarity or dissimilarity between two instances, which should satisfy the non-negativity, symmetry, and triangle inequality. Then a metric learning algorithm can be justified as a parameterized optimization process that minimizes a loss function over pairs or triplets of instances.
\[
min_\theta \sum_{x_i \sim x_j } d^2(x_i,x_j) + \lambda \sum_{x_i \not\sim x_j } d^2(x_i,x_j)
\]
which, in layman's terms, encourages the embedding of similar instances to be close together in the feature space.

We use a bottom-up approach to train trunk and embedder models for deep metric learning. First, individual segments are trained separately, with the weights of the resulting trunk and embedder models saved. Next, each individual segment is paired with another individual segment, with the weights from their previous trained models averaged together and set as the default weights before training the new trunk and embedder. This continues until all segments have been combined to reconstruct the original image.

Deep metric learning begins by obtaining the embeddings and labels for the training data. Specifically, the embeddings are gathered by running the trunk model, which is typically a CNN such as Resnet, then feeding the output of the trunk model to the embedder model to reduce the dimensions further. The embedder model is an MLP model that reduces the trunk output to a small number of dimensions via a single hidden layer.

The main goal of deep metric learning is to create embeddings where similar images are closer together and dissimilar images are farther apart. To do so, the loss function while training the models is based on the distance between the embeddings. Since this distance is represented by a matrix, mining is used to find the hardest pairs of training samples to target those in the next training epoch in the loss function.

In terms of explainability, we focus on explaining the final, global decision $z$ using local decisions at each segment. Starting with the individual segments, a local decision $y_i$ is generated. As segments are combined, the local decisions are also combined into a new local decision, such as $y_3 = y_1 + y_2$. This repeats until all local decisions have been combined to create the global decision, $z = \sum_i^N y_i$.


To get the best results, we allow for different loss functions to be used. By default, the models are trained with triplet margin loss with a margin of 0.1 as the loss function, which is calculated as
\begin{align*}
    \mathcal{L}=|d_{ap}-d_{an}+m|
\end{align*}
where $d_{ap}$ is the distance between the positive anchor and the sample triplets, $d_{an}$ is the distance between the negative anchor and sample triplets, and $m$ is the margin of 0.1 \cite{schroff2015facenet}. 


Algorithm \ref{alg:1} contains the basic training loop for training all segment models. Given the list of individual and combined segments, it iterates through all the segments starting with the individual ones. For each segment, it creates the dataset with the appropriate segmented images. If the current segment is a combination of two prior segments, then the models are loaded with the weighted average of the individual segment weights that were previously trained. Once the dataset and the model are set up, the model is then trained using a miner to gather hard pairs to optimize via the loss function. This repeats until all training epochs are completed.

\begin{algorithm}
\caption{Basic Training Loop}\label{alg:1}
    \KwData{segments}
    \KwResult{trained segment models}
    \For{segment in segments}{
        $dataset = \texttt{create\_dataset}(segment)$\;
        \If{segment is combined}{
            $\texttt{load\_weights\_averaged}(\texttt{model})$\;
        }
        \For{data, labels in dataset}{
            $x = \texttt{model}(data)$\;
            $pairs = \texttt{miner}(x,\,labels)$\;
            $loss = \mathcal{L}(x,\,labels,\,pairs)$\;
        }
        $\texttt{save\_weights}(\texttt{model})$\;
    }
\end{algorithm}

\subsection{Metric Tree Construction}
\label{sec:tree}
To aid in the explanation, we construct a visualization of the similarity between two images at each level within the semantic reconstruction. The saved weights for the best accuracy at each level are reloaded to get the embeddings for the two query images. The embeddings are then compared using several different distance and similarity metrics. Both metrics report results in the range of 0 to 1. For a distance metric, a value near zero indicates that there is a small amount of distance between the two queried images. For a similarity metric, this is inverted, where a value near one indicates that there is a high similarity between the two queried images. We use the distance metric as our main explainability component.

By default, tree composition uses Signal-to-Noise Ratio (SNR) as the primary distance metric \cite{yuan2019signaltonoise}. The SNR distance is calculated as
\begin{align*}
    d_{SNR} = \sum_i^N \frac{\sigma^2(x[i] - y[i])}{\sigma^2(x[i])}
\end{align*}
where $x$ and $y$ are the $N$-dimension embeddings of the two images being compared.

An inference model consisting of the trunk and embedder models with the saved weights is used at each specific segment. This inference model is utilized to get the weights for any queried image of a relevant image to what the inference model is expecting. For example, the ``hair" inference model will be used to query only hair segment images.
\section{Experiments}
\label{sec:result}
The following sections demonstrate the effectiveness of our hierarchical explainable metric learning. We first describe the experimental setup. Next, we present three case studies.

\subsection{Experimental Setup}
All experiments were run on Ubuntu 22.04 with the following configuration: Intel Core i9-13900K CPU, NVIDIA RTX 4090 GPU, and 64 GB RAM. We used Python v3.11.4, PyTorch 2.0.1, and PyTorch Metric Learning \cite{musgrave2020pytorch} in our code. For datasets, we use CelebA \cite{liu2015deep}, SceneParse150 derived from ADE20k \cite{zhou2016semantic, zhou2017scene}, and Human Parsing \cite{liang2015deep}. We have provided experimental results using 7 different loss functions: Triplet Margin, Angular, Multiple Similarity, NTXent, Signal-to-Noise Ratio, Proxy Anchor, and Sub-Center Arc Face \cite{schroff2015facenet, deng2020subcenter, chen2020simple, he2020momentum, kim2020proxy, wang2017deep, wang2019multisimilarity, yuan2019signaltonoise, oord2019representation}. 

We compared our results to previous work, namely traditional deep metric learning and AVSL. Our methodology achieved comparable accuracy with significantly less memory footprint compared to AVSL. We provide a visualization of the similarity between the semantic segments of an image and how they are similar as they are combined back into the original image. To measure the performance of the model, we use Precision@K as our main metric. Most notably, we use Precision@1, Precision@2, and Precision@8.


\begin{table*}[h]
    \centering
        \caption{Precision@1, Precision@2, and Precision@8 results for the CelebA, Human Parsing (HP), and SceneParse150 (SP150) datasets for different models. We compare our approach (last seven rows) with previous methods of traditional deep metric learning (first two rows) and AVSL (third row). The models are separated by the loss function used.}
        \vspace{-0.1in}
    \begin{tabular}{lccccccccc} \toprule
        \multicolumn{1}{c}{} & \multicolumn{3}{c}{\textbf{CelebA}} & \multicolumn{3}{c}{\textbf{HP}}& \multicolumn{3}{c}{\textbf{SP150}}\\
        \cmidrule(rl){2-4} \cmidrule(rl){5-7} \cmidrule(rl){8-10}
        \textbf{Model} & P@1 & P@2 & P@8& P@1 & P@2 & P@8& P@1 & P@2 & P@8\\
        \midrule
        Triplet \cite{schroff2015facenet} & 88.1 & 89.0 & 88.6 & 65.3 & 65.4 & 65.0 & 88.4 & 87.9 & 87.6\\
        ProxyAnchor \cite{kim2020proxy} & 80.2 & 79.7 & 78.5 & 62.4 & 61.8 & 61.2 & 83.1 & 82.3 & 81.6 \\
        \midrule
        ProxyAnchor-AVSL \cite{kim2020proxy, zhang2022attributable} & 88.5 & 88.3 & 87.3 & 72.8 & 71.6 & 70.5 & 89.5 & 88.8 & 88.8 \\
        \midrule
        Triplet-HEML \cite{schroff2015facenet} & 88.2 & 88.5 & 88.0 & 66.8 & 66.1 & 65.9 & 87.5 & 87.4 & 87.6\\ 
        Angular-HEML \cite{wang2017deep} & 80.5 & 80.6 & 80.4 & 63.3 & 63.6 & 63.3 & 85.1 & 84.4 & 84.6 \\
        MultiSim-HEML \cite{wang2019multisimilarity} & 87.5 & 87.5 & 87.8 & 65.8 & 64.6 & 64.1 & 88.3 & 88.0 & 87.8 \\
        NTXent-HEML \cite{oord2019representation, chen2020simple, he2020momentum} & 89.3 & 89.7 & 89.4 & 66.7 & 66.6 & 66.0 & 87.6 & 87.4 & 87.9 \\
        SNR-HEML \cite{yuan2019signaltonoise} & 90.0 & 89.8 & 89.5 & 68.1 & 67.5 & 66.7 & 87.9 & 87.7 & 88.1\\
        ProxyAnchor-HEML \cite{kim2020proxy} & 72.1 & 72.4 & 72.8 & 62.7 & 61.6 & 61.4 & 84.3 & 83.0 & 82.4 \\
        SubCenter-HEML \cite{deng2020subcenter} & 89.3 & 88.4 & 88.5 & 66.4 & 66.2 & 65.9 & 85.6 & 85.3 & 85.6 \\
        \bottomrule
    \end{tabular}
    \label{tbl:loss}
\end{table*}

Our methodology requires less GPU memory than AVSL and the same GPU memory as traditional. However, this comes with the tradeoff of longer run times and increased CPU utilization. HEML will take approximately 2-3 GB of GPU memory throughout its training time, whereas AVSL will take between 18-20 GB. Traditional deep metric learning uses approximately 2-3 GB. Due to training models at each level, extra CPU and disk utilization is required to set up the current segment data loader and save the resulting trained models. However, this increased overhead is not too large compared to the overhead for similar operations with the prior works.

\subsection{Case Study 1: CelebA}
\begin{figure}[h]
    \centering
    \begin{tikzpicture}
        \node[inner sep=0pt] (final) at (0,0) {\includegraphics[scale=0.4]{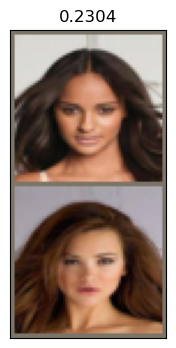}};
        \node[inner sep=0pt] (l1) at (-1, -4) {\includegraphics[scale=0.4]{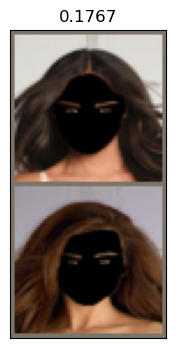}};
        \node[inner sep=0pt] (l2) at (1, -4) {\includegraphics[scale=0.4]{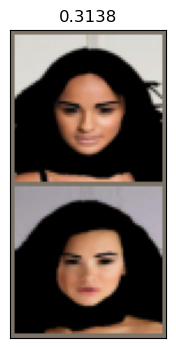}};
        \node[inner sep=0pt] (l3) at (-3, -8) {\includegraphics[scale=0.4]{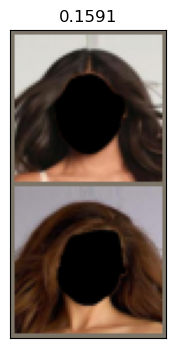}};
        \node[inner sep=0pt] (l4) at (-1, -8) {\includegraphics[scale=0.4]{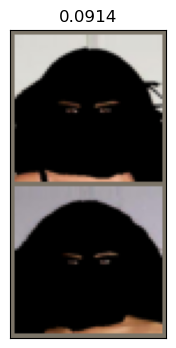}};
        \node[inner sep=0pt] (l5) at (1, -8) {\includegraphics[scale=0.4]{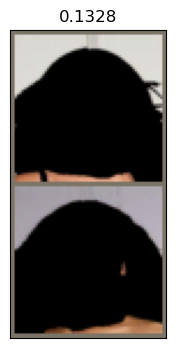}};
        \node[inner sep=0pt] (l6) at (3, -8) {\includegraphics[scale=0.4]{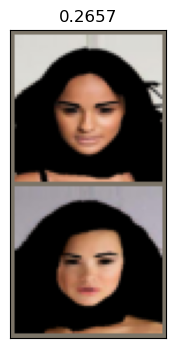}};

        \draw[->] (final.south) to[|-|] (l1.north);
        \draw[->] (final.south) to[|-|] (l2.north);
        \draw[->] (l1.south) to[|-|] (l3.north);
        \draw[->] (l1.south) -- (l4.north);
        \draw[->] (l2.south) -- (l5.north);
        \draw[->] (l2.south) to[|-|] (l6.north);
    \end{tikzpicture}
    \caption{Hierarchical visualization of two sample comparisons from the CelebA dataset. The value above each image comparison indicates the SNR distance between the two images. A value closer to 0 indicates the images are similar, and a value closer to 1 indicates the images are dissimilar. The distances are measured via an inference model using the trained Triplet-HEML model for CelebA. Segments from left to right on the bottom row: neck, cloth, hair, hat; eyes, brows; ears, ear\_r, eye\_g; nose, skin, lips.}
    \label{fig:celeb}
\end{figure}

The CelebA dataset contains images of faces. We use a subset of 5,000 training images and 1,000 validation images from the original dataset. Figure \ref{fig:celeb} showcases the hierarchical visualization of two sample images after training Triplet-HEML. For each image pair based on segmentation, the SNR distance is calculated and displayed. With this, we can get a clear picture of what segments are contributing to the distance for the fully reconstructed image. In this example, the neck, hair, eyes, and brow segments contribute more to the overall similarity of the two images compared to the skin, ear, nose, and lip segments.



When compared against traditional deep metric learning on CelebA, the normal method gains a Precision@1 of 88.1\% on the validation set using the ``Male" label for classification, as shown in Table \ref{tbl:loss}. The top level of our method gains a Precision@1 of 88.2\% under the same constraints. Our methodology does introduce some memory and timing overhead, however, we feel that the tradeoff by introducing more explainable distance metrics is a valid tradeoff. The training time for HEML was approximately 5 minutes, whereas the traditional was approximately 3 minutes. The memory usage is comparable with traditional.

Table \ref{tbl:1} displays the accuracy values from each level of segment combinations. For example, the ``Neck" segment has a Precision@1 of 78.5\% and the ``Cloth" segment 79.6\%, but when trained using these two segments combined, the resulting Precision@1 is 80.1\%.

\begin{table*}[h]
    \centering
        \caption{Performance comparison of HEML, traditional, and AVSL on CelebA, Human Parsing (HP), and SceneParse150 (SP150). Exact performance varies within ranges provided across multiple runs.}
        \vspace{-0.1in}
    \begin{tabular}{lcccccc}
        \toprule
        & \multicolumn{2}{c}{\textbf{CelebA}} & \multicolumn{2}{c}{\textbf{HP}} & \multicolumn{2}{c}{\textbf{SP150}}\\
        \cmidrule(rl){2-3} \cmidrule(rl){4-5} \cmidrule(rl){6-7}
        \textbf{Method} & \textbf{Training Time} & \textbf{GPU memory} & \textbf{Training Time} & \textbf{GPU memory}& \textbf{Training Time} & \textbf{GPU memory}\\
        \midrule
        HEML & 4-6 min & 2-3 GB & 17-19 min & 2-3 GB & 5-6 min & 2-3 GB \\
        Traditional & 3 min & 2-3 GB & 10 min & 2-3 GB & 2 min & 2-3 GB\\
        AVSL & 2-3 min & 18-20 GB & 3-4 min & 18-20 GB & 3-4 min & 18-20 GB\\
        \bottomrule
    \end{tabular}
    \label{tab:perf}
\end{table*}

\begin{table}[h]
    \centering
        \caption{Precision@1, Precision@2, and Precision@8 for each segment and segment combination for CelebA using Triplet-HEML. The ``final" segment indicates a fully reconstructed image.}
        \vspace{-0.1in}
    \begin{tabular}{lccc}
    \toprule
    \textbf{Segment} & \textbf{P@1} & \textbf{P@2} & \textbf{P@8} \\
    \midrule
    Neck & 78.5 & 79.6 & 80.4 \\
    Cloth & 79.6 & 79.6 & 79.7 \\
    Hair & 83.9 & 83.6 & 82.9 \\
    Hat & 80.4 & 79.1 & 77.9 \\
    L\_eye & 79.5 & 79.4 & 79.3 \\
    R\_eye & 79.6 & 79.6 & 78.9 \\
    L\_brow & 77.8 & 77.9 & 78.1 \\
    R\_brow & 77.6 & 78.1 & 78.3 \\
    L\_ear & 75.0 & 75.2 & 76.1 \\
    R\_ear & 76.2 & 76.2 & 76.2 \\
    Ear\_r & 77.9 & 77.7 & 77.9 \\
    Eye\_g & 77.8 & 78.6 & 78.5 \\
    Nose & 79.8 & 79.3 & 78.5 \\
    Skin & 85.7 & 85.3 & 85.6 \\
    U\_lip & 77.1 & 77.2 & 76.8 \\ 
    L\_lip & 77.9 & 77.3 & 77.0 \\
    \midrule
    $a=$ Neck + Cloth & 80.1 & 80.2 & 80.5 \\
    $b=$ Hair + Hat & 81.6 & 82.0 & 82.4 \\
    $c=$ L\_eye + R\_eye & 79.3 & 78.9 & 78.9 \\
    $d=$ L\_brow + R\_brow & 79.6 & 79.8 & 79.6 \\
    $e=$ L\_ear + R\_ear & 77.6 & 76.6 & 77.7 \\
    $f=$ Ear\_r + Eye\_g & 78.2 & 78.6 & 79.5 \\
    $g=$ Nose + Skin & 86.1 & 86.7 & 86.6 \\
    $h=$ U\_lip + L\_lip & 75.7 & 76.2 & 76.2 \\
    \midrule
    $m=a+b$ & 80.5 & 81.4 & 81.3 \\
    $n=c+d$ & 83.1 & 83.2 & 81.9 \\
    $o=e+f$ & 76.0 & 77.0 & 77.7 \\
    $p=g+h$ & 87.0 & 86.8 & 87.1 \\
    \midrule
    $x=m+n$ & 86.5 & 86.2 & 86.5 \\
    $y = o + p$ & 86.4 & 86.2 & 86.3 \\
    \midrule
    final $=x+y$ & 88.2 & 88.5 & 88.0\\
    \bottomrule
    \end{tabular}
    \label{tbl:1}
\end{table}

\begin{figure*}
    \centering
    \begin{tikzpicture}
        \draw[<->] (0,0) node [above right, font=\scriptsize] {more similar} -- (15,0) node [above left, font=\scriptsize] {less similar};
    \end{tikzpicture}\\
    \includegraphics[scale=0.28]{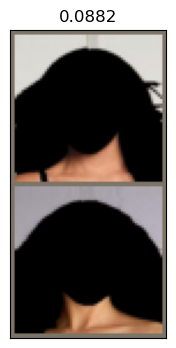}
    \includegraphics[scale=0.28]{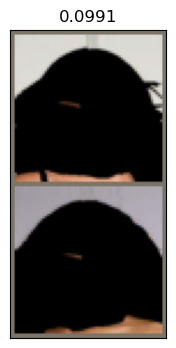}
    \includegraphics[scale=0.28]{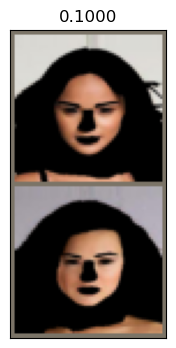}
    \includegraphics[scale=0.28]{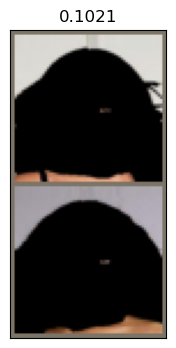}
    \includegraphics[scale=0.28]{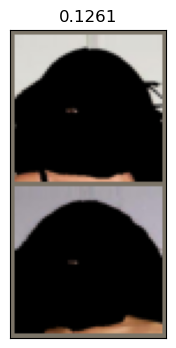}
    \includegraphics[scale=0.28]{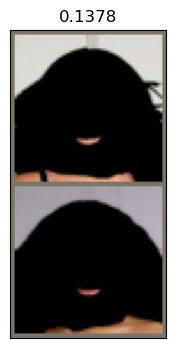}
    \includegraphics[scale=0.28]{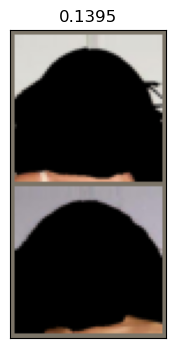}
    \includegraphics[scale=0.28]{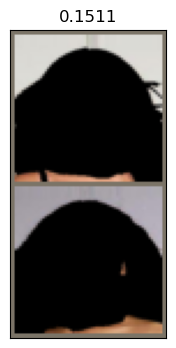}
    \includegraphics[scale=0.28]{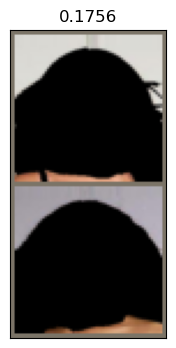}
    \includegraphics[scale=0.28]{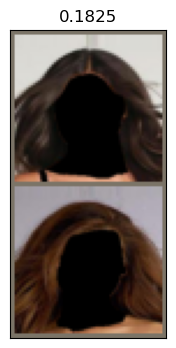}
    \includegraphics[scale=0.28]{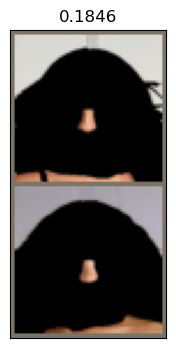}
    \includegraphics[scale=0.28]{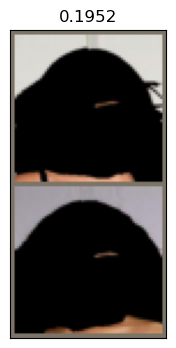}
    \includegraphics[scale=0.28]{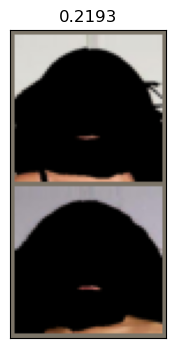}
    \caption{Individual segments and SNR distance between, as indicated by value above each image. From left to right: neck, r\_brow, skin, l\_eye, r\_eye, l\_lip, cloth, l\_ear, r\_ear, hair, nose, l\_brow, u\_lip.}
    \label{fig:enter-label}
\end{figure*}

\subsection{Case Study 2: Human Parsing}
For the Human Parsing dataset, the Precision@1 of Triplet-HEML using 16 segments is 66.8\% and it takes approximately 17 minutes to train. In comparison, traditional deep metric learning is 65.3\% with 10-minute training. Memory usage is comparable between HEML and traditional.

Since Human Parsing does not have labels for classification by default, we create a binary labeling system based on the available segments. Suppose an image has a ``Dress" segment. In that case, it is labeled with a value of 1, and 0 otherwise. This labeling system is also loosely aligned with CelebA's labeling system using the ``Male" attribute for classification. The labeling scheme may be a cause of the low Precision@K metrics that all models, including those from prior works, have. Creating classification labels for a dataset not intended for classification may result in errors or unclear label boundaries.

\subsection{Case Study 3: SceneParse150}
The SceneParse150 dataset contains 20,210 training images and 2,000 validation images consisting of different visual scenes such as buildings and landscapes. This dataset achieved a Precision@1 of 87.5\% using Triplet-HEML and 88.4\% with Triplet traditional deep metric learning. Training with HEML takes approximately 5 minutes compared to traditional's 2 minutes. Memory usage is comparable between traditional and HEML.

As this dataset contains 150 unique segments, we chose to focus on some of the more prevalent segments in the images. The segments chosen are wall, building, sky, tree, grass, plant, house, and path. We create binary classification labels for the Precision@K metrics using the ``building" segment, similar to the ``Dress" segment with Human Parsing.

Table~\ref{tab:perf} provides a summary of all three case studies. For training times, HEML takes approximately 2 to 3 GB to train various models, while state-of-the-art (AVSL) requires 18 to 20 GB. Although traditional has comparable memory overhead, it does not provide explainability. Overall, our proposed framework provides explainability with reasonable memory requirements.

\section{Conclusion}
\label{sec:conclusion}
Explainable deep metric learning is popular due to its ability to provide human-understandable similarity between two images. However, the existing approaches can introduce unacceptable memory overhead due to saliency map construction for each image. We presented an efficient explainable metric learning framework, Hierarchical Explainable Machine Learning (HEML), for explainable similarity metrics between two images at different semantic segmentation levels. Our approach provides a lightweight solution due to the one-time construction of segment samples. Moreover, our metric tree construction provides inherent metric explainability. Extensive experimental evaluation using three case studies demonstrated that our approach provides comparable accuracy with a drastic reduction in memory requirement compared to state-of-the-art explainable metric learning methods. 
\section*{Acknowledgments}
This work was partially supported by a Semiconductor Research Corporation (SRC) grant 2023-AI-3155.

\bibliographystyle{IEEEtran}
\bibliography{IEEEabrv,main.bib}

\end{document}